\documentclass[11pt,a4paper]{article}
\usepackage[hyperref]{conll2018}
\usepackage{times}
\usepackage[caption=false]{subfig}
\usepackage{times,graphicx}
\usepackage{latexsym,booktabs,multicol,multirow}
\usepackage{amsmath}
\usepackage{latexsym}

\usepackage{url}
\usepackage{pifont}% http://ctan.org/pkg/pifont
\newcommand{\cmark}{\ding{51}}%
\newcommand{\xmark}{\ding{55}}%
\aclfinalcopy % Uncomment this line for the final submission

\title{Generalizing Procrustes Analysis for Better Bilingual Dictionary Induction}

\author  
  {
	\begin{tabular}{cccc}
	Yova Kementchedjhieva$^{\Diamond}$ & Sebastian Ruder$^{\spadesuit\clubsuit}$ & Ryan Cotterell$^\heartsuit$ & Anders S{\o}gaard$^{\Diamond}$
	\end{tabular}
	\\
    $^\Diamond$University of Copenhagen, Copenhagen, Denmark\\
    $^\spadesuit$Insight Research Centre, National University of Ireland, Galway, Ireland\\
    $^\clubsuit$Aylien Ltd., Dublin, Ireland\\
    $^\heartsuit$University of Cambridge, Cambridge, UK\\
	{\tt \small{\{yova|soegaard\}@di.ku.dk,sebastian@ruder.io,ryan.cotterell@gmail.com,}}
}

\date{}

\begin{document}
\maketitle
\begin{abstract}
Most recent approaches to bilingual dictionary induction find a linear alignment between the word vector spaces of two languages. We show that projecting the two languages onto a third, latent space, rather than directly onto each other, while  equivalent in terms of expressivity, makes it easier to learn approximate alignments. %expressiveness of the alignment. 
Our modified approach also allows for supporting languages to be included in the alignment process, to obtain an even better performance in low resource settings.
%We show that weakly-supervised bilingual dictionary induction from word vector spaces using  adversarial networks improves from projecting into a third, common space. While the third space does not increase the expressiveness of our alignment model, fixing the target language embedding can be a serious bottleneck for learning a linear transformation between approximately isomorphic vector spaces. We show that further bias is reduced by introducing a third, supporting language, and projecting all three languages into a fourth, common space. 

%Cross-lingual learning has been observed to be more robust when transfer is regularized by interpolating from several source languages. Unsupervised bilingual dictionary induction (UBDI) from monolingual embeddings is sensitive to linguistic differences between the two languages, domain mismatches, and small biases in the monolingual embeddings \cite{Soegaard:ea:18}. We present a simple extension to UBDI that learns to map target language vectors into a {\em multisource} vector space. This leads to better performance across all the languages we consider; but more importantly, it makes UBDI more robust, i.e., less vulnerable to the instability of GANs.  
\end{abstract}

\section{Introduction}
Several papers recently demonstrated the potential of very weakly supervised or entirely unsupervised approaches to bilingual dictionary induction (BDI) \cite{Barone2016,Artetxe2017,Zhang2017c,Conneau2018,Søgaard2018}, the task of identifying translational equivalents across two languages. These approaches cast BDI as a problem of aligning monolingual word embeddings. Pairs of monolingual word vector spaces can be aligned without any explicit cross-lingual supervision, solely based on their distributional properties (for an adversarial approach, see \citet{Conneau2018}). Alternatively, weak supervision can be provided in the form of numerals \cite{Artetxe2017} or identically spelled words \cite{Søgaard2018}. Successful unsupervised or weakly supervised alignment of word vector spaces would remove much of the data bottleneck for machine translation and push horizons for cross-lingual learning ~\cite{Ruder2018}.
% One of the key extrinsic indicators of the quality of cross-lingual alignment of vector spaces is performance on the task of bilingual dictionary induction (BDI), i.e. the identification of translational equivalents across two languages. BDI itself is also a key step in unsupervised machine translation, which starts off by identifying word-for-word translations \cite{Lample2018}.

In addition to an unsupervised approach to aligning monolingual word embedding spaces with adversarial training, \citet{Conneau2018} %, a method for learning Multilingual Unsupervised and Supervised Embeddings
present a supervised alignment algorithm that assumes a gold-standard seed dictionary and performs Procrustes Analysis \cite{schonemann1966procrustes}. \citet{Søgaard2018} show that this approach, weakly supervised with a dictionary seed of \textit{cross-lingual homographs}, i.e. words with identical spelling across source and target language, is superior to the completely unsupervised approach. We therefore focus on weakly-supervised Procrustes Analysis (PA) for BDI here. 

%This alignment takes the form of a linear matrix that maps the source space into the target space in accordance with the previous literature \cite{Mikolov2013e,Xing2015}. BDI is then performed by finding the nearest neighbor of a source word in the target space and assuming the two to be translational equivalents.\par 

The implementation of PA in \newcite{Conneau2018} yields notable improvements over earlier work on BDI, even though it learns a simple linear transform of the source language space into the target language space. 
Seminal work in supervised alignment of word vector spaces indeed reported superior performance with linear models as compared to non-linear neural approaches \cite{Mikolov2013e}. 
The relative success of the simple linear approach can be explained in terms of isomorphism across monolingual semantic spaces,\footnote{Two vector spaces are isomorphic if there is an invertible linear transformation from one to the other.} an idea that receives support from cognitive science \cite{Youn2016}. Word vector spaces are not \textit{perfectly} isomorphic, however, as shown by \citet{Søgaard2018}, who use a Laplacian graph similarity metric to measure this property. In this work, we show that projecting both source and target vector spaces into a {\em third}~space \cite{Faruqui2014}, using a variant of PA known as Generalized Procrustes Analysis \cite{gower1975generalized}, makes it easier to learn the alignment between two word vector spaces, as compared to the single linear transform used in \citet{Conneau2018}.

%Some linear alignment models thus experiment with projecting both source and target vector spaces into a {\em third}~space \cite{Faruqui2014}.

%In the context of perfectly isomorphic word vector spaces, the introduction of a third space would be redundant: if there is a linear transform from a word vector space $E$ to a third space $S$, and a linear transform from a word vector space $F$ to $S$, there is also a linear transform from $E$ to $F$. 

%One limitation of projecting into the target language is that while such an approach can map multiple languages into this target language vector space, the alignments will be learned independently of each other, and 
%can only be used for two languages at a time, while often in cross-lingual transfer learning a larger number of languages may be involved. Another more theoretical shortcoming is that MUSE learns a single matrix to capture the direct alignment between two language spaces, which may be overly restrictive given that monolingual vector spaces have been shown not to be approximately isomorphic in many cases \cite{Søgaard2018}. Introducing a third, latent space, which each language space gets mapped to independently, offers potential for better performance.\par

\paragraph{Contributions} We show that %an alternative form of Procrustes, known as 
Generalized Procrustes Analysis (GPA) \cite{gower1975generalized}, a method that maps two vector spaces into a third, latent space, is superior to PA for BDI, % lends itself nicely to the task of learning an alignment between multiple languages, with a minor drop in pairwise accuracy compared to the bilingual setup. Moreover, GPA yields an improvement in performance for BDI over simple Procrustes, 
e.g., improving the state-of-the-art on the widely used English-Italian dataset \cite{Dinu2015} from a P@1 score of 66.2\% to 67.6\%. We compare GPA to PA on aligning English with five languages representing different language families (Arabic, German, Spanish, Finnish, and Russian), showing that GPA consistently outperforms PA. GPA also allows for the use of additional support languages, aligning three or more languages at a time, which can boost performance even further. We present experiments with multi-source GPA on an additional five low-resource languages from the same language families (Hebrew, Afrikaans, Occitan, Estonian, and Bosnian), using their bigger counterpart as a support language. Our code is publicly available.\footnote{\url{https://github.com/YovaKem/generalized-procrustes-MUSE}} \par

%\section{Related work}

%\paragraph{Bilingual embeddings} 

%\paragraph{Multilingual embeddings}
\section{Procrustes Analysis}

Procrustes Analysis is a graph matching algorithm, used in most mapping-based approaches to BDI \cite{Ruder2018}. 
%It is visualized conceptually in Figure~\ref{fig:pa_gpa} (a). 
Given two graphs, $E$ and $F$, Procrustes finds the linear transformation $T$ that minimizes the following objective:
\begin{equation} \label{eq:procrustes}
\arg\min_T\ ||TE - F||^2
\end{equation}
thus minimizing the trace between each two corresponding rows of the transformed space $TE$ and $F$. We build $E$ and $F$ based on a seed dictionary of $N$ entries, such that each pair of corresponding rows in $E$ and $F$, $(e_n, f_n)$ for $n=1,\ldots,N$ consists of the embeddings of a translational pair of words. 
%, where a matrix is learned that maps the source language space ($X_{1}$) onto the target language space ($X_2$). Since $X_1$ and $X_2$ need to have corresponding rows, MUSE initializes Procrustes with a seed dictionary. The word pairs in this dictionary serve as anchor points for the complete embeddings spaces.  \par
In order to preserve the monolingual quality of the transformed embeddings, it is beneficial to use an orthogonal matrix $T$ for cross-lingual mapping purposes \cite{Xing2015,Artetxe2017}.\footnote{Recently, \citet{doval2018improving} showed that the monolingual quality of embeddings need not suffer from a transformation guided by cross-lingual alignment, but their method still relies on an initial alignment obtained e.g. with Procrustes analysis, as described here.} Conveniently, the orthogonal Procrustes problem has an analytical solution, based on Singular Value Decomposition (SVD): 
\begin{equation}
\begin{split}
F^\top E & = U\Sigma V^\top \\
T & = VU^\top
\end{split}
\end{equation}

\begin{figure}
\centering
\subfloat[Procrustes Analysis]{%
  \includegraphics[clip,width=0.7\columnwidth]{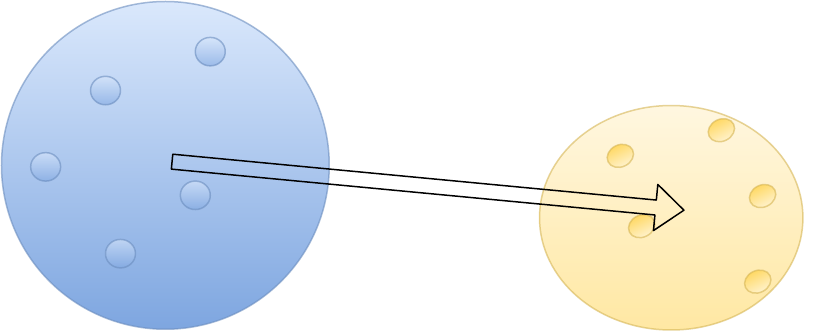}%
}

\subfloat[Generalized Procrustes Analysis]{%
  \includegraphics[clip,width=0.7\columnwidth]{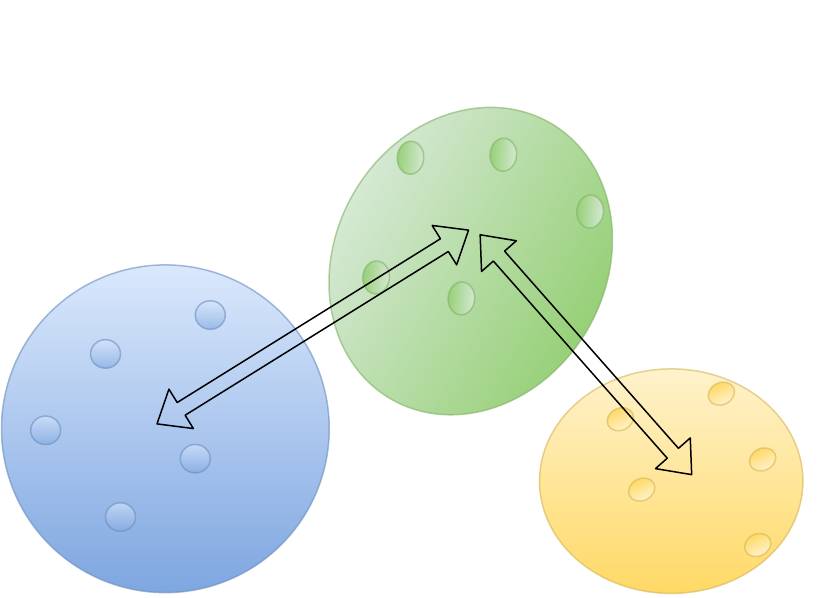}%
}
\caption{Visualization of the difference between PA, which maps the source space directly onto the target space, and GPA, which aligns both source and target spaces with a third, latent space, constructed by averaging over the two language spaces.}
\label{fig:pa_gpa}
\end{figure}

%PA can be applied iteratively by extracting a new dictionary seed from each updated alignment of the vector spaces. 

\section{Generalized Procrustes Analysis}
Generalized Procrustes Analysis  \cite{gower1975generalized} is a natural extension of PA that aligns $k$ vector spaces at a time. 
%Figure~\ref{fig:pa_gpa} (b) visualizes GPA with $k=2$. 
Given embedding spaces $E_1,\ldots, E_{k}$, GPA minimizes the following objective:

\begin{equation}\label{eq:gprocrustes}
\arg\min_{\{T_1,\ldots,T_k\}}\sum_{i<j}^k ||T_i E_i - T_j E_j||^2
\end{equation}

%\begin{equation}
%\sum_{i<j}^k \mathrm{Tr} (X_{i}T_i - X_{j}T_j)
%\end{equation}
%The analytical solution is not directly applicable here, since the objective cannot be represented as a one-sided transformation (unless we select an anchor space and transform all remaining languages to match it, which is suboptimal). One way around this obstacle is to 

For an analytical solution to GPA, we compute %a \textit{group average configuration} $G$, defined as 
the average of the  embedding matrices $E_{1\ldots k}$ after transformation by $T_{1\ldots k}$:
\begin{equation}\label{eq:compute_g}
G={k^{-1}}\sum_{i=1}^k E_i T_i
\end{equation}
thus obtaining a latent space, $G$, which captures properties of each of $E_{1\ldots k}$, and potentially additional properties emerging from the combination of the spaces. On the very first iteration, prior to having any estimates of $T_{1\ldots k}$, we set $G=E_i$ for a random $i$. 
The new values of $T_{1\ldots k}$ are then obtained as:
\begin{equation}\label{eq:update_t}
\begin{split}
G^\top E_i & = U\Sigma V^\top\\
T_i & = VU^\top \text{ for } i \text{ in } 1\ldots k
\end{split}
\end{equation}
Since $G$ is dependent on $T_{1\ldots k}$ (see Eq.\ref{eq:compute_g}), the solution of GPA cannot be obtained in a single step (as is the case with PA), but rather requires that we loop over subsequent updates of $G$ (Eq.\ref{eq:compute_g}) and $T_{1\ldots k}$ (Eq.\ref{eq:update_t}) for a fixed number of steps or until satisfactory convergence. We observed little improvement when performing more than 100 updates, so we fixed that as the number of updates. 

Notice that for $k=2$ and with the orthogonality constraint in place, the objective for Generalized Procrustes Analysis (Eq.~\ref{eq:gprocrustes}) reduces to that for simple Procrustes (Eq.~\ref{eq:procrustes}):
\begin{equation}
\begin{split}
\arg\min_{\{T_1,T_2\}} ||T_1 E_1 - T_2 E_2||^2 \\
=\arg\min_{T} ||T E_1 -E_2||^2 \\
\text{ where } T=T_1T_2^T
\end{split}
\end{equation}
Here $T$ itself is also orthogonal. Yet, the solution found with GPA may differ from the one found with simple Procrustes: the former maps $E_1$ and $E_2$ onto a third space, $G$, which is the average of the two spaces, instead of mapping $E_1$ directly onto $E_2$. To understand the consequences of this difference, consider a single step of the GPA algorithm where after updating $G$ according to Eq.\ref{eq:compute_g} we are recomputing $T_1$ using SVD. Due to the fact that $G$ is partly based on $E_1$, these two spaces are bound to be more similar to each other than $E_1$ and $E_2$ are.\footnote{A theoretical exception being the case there $E_1$ and $E_2$ are identical.} Finding a good mapping between $E_1$ and $G$,  i.e. a good setting of $T_1$, should therefore be easier than finding a good mapping from $E_1$ to $E_2$ directly. In this sense, by mapping $E_1$ onto $G$, rather than onto $E_2$ (as PA would do), we are solving an easier problem and reducing the chance of a poor solution.

\begin{table}
\resizebox{\linewidth}{!}{
\begin{tabular}{l|l|l|l|l|l}
\toprule
High-resource&{\sc ar}&{\sc de}&{\sc es}&{\sc fi}&{\sc ru}\\
\midrule
&575k&2,183k&1,412k&437k&1,474k\\
\midrule\midrule
Low-resource&{\sc he}&{\sc af}&{\sc oc}&{\sc et}&{\sc bs}\\
\midrule
&224k&49k&84k&175k&77k\\
\bottomrule
\end{tabular}
}
\caption{Statistics for Wikipedia corpora.}
\label{tab:stats}
\end{table}

\section{Experiments}

\begin{table*}
\resizebox{\textwidth}{!}{
\begin{tabular}{l|ll|ll|ll|ll|ll|ll}
\toprule
& \multicolumn{2}{c|}{\sc ar}& \multicolumn{2}{c|}{\sc de}& \multicolumn{2}{c|}{\sc es}& \multicolumn{2}{c|}{\sc fi}& \multicolumn{2}{c|}{\sc ru}& \multicolumn{2}{c}{Ave}\\
& $k=1$ & $k=10$& $k=1$ & $k=10$& $k=1$ & $k=10$& $k=1$ & $k=10$& $k=1$ & $k=10$& $k=1$ & $k=10$\\
\midrule
PA&34.73&61.87&73.67&91.73&81.67&92.93&45.33&75.53&47.00&79.00&56.48&80.21\\
\midrule
GPA&{\bf 35.33}&\textbf{64.27}&{\bf 74.40}&\textbf{91.93}&{\bf 81.93}&\textbf{93.53}&{\bf 47.87}&\textbf{76.87}&{\bf 48.27}&\textbf{79.13}&{\bf 57.56}&\textbf{81.15}\\
\bottomrule
\end{tabular}
}
\caption{Bilingual dictionary induction performance, measured in P@k, of PA and GPA across five language pairs.}
\label{tab:pa-gpa-comparison}
\end{table*}

In our experiments, we generally use the same hyper-parameters as used in \citet{Conneau2018}, unless otherwise stated. When extracting dictionaries for the bootstrapping procedure, we use cross-domain local scaling (CSLS, see \citet{Conneau2018} for details) as a metric for ranking candidate translation pairs, and we only use the ones that rank higher than 15,000. We do not put any restrictions on the initial seed dictionaries, based on cross-lingual homographs: those vary considerably in size, from 17,012 for Hebrew to 85,912 for Spanish. 
Instead of doing a single training epoch, however, we run PA and GPA with early stopping, until five epochs of no improvement in the validation criterion as used in \citet{Conneau2018}, i.e. the average cosine similarity between the top 10,000 most frequent words in the source language and their candidate translations as induced with CSLS.
Our metric is Precision at k$\times 100$ (P@k), i.e. percentage of correct translations retrieved among the $k$ nearest neighbor of the source words in the test set \cite{Conneau2018}. Unless stated otherwise, experiments were carried out using the publicly available pre-trained fastText embeddings, trained on Wikipedia data,\footnote{\url{https://github.com/facebookresearch/fastText}} and bilingual dictionaries---consisting of 5000 and 1500 unique word pairs for training and testing, respectively---provided by \citet{Conneau2018}\footnote{\url{https://github.com/facebookresearch/MUSE}}.    

\subsection{Comparison of PA and GPA}
\paragraph{High resource setting}
We first present a direct comparison of PA and GPA on BDI from English to five fairly high-resource languages: Arabic, Finnish, German, Russian, and Spanish. The Wikipedia corpus sizes for these languages are reported in Table~\ref{tab:stats}. \textbf{Results} are listed in Table \ref{tab:pa-gpa-comparison}. GPA improves over PA consistently for all five languages. Most notably, for Finnish it scores 2.5\% higher than PA. 

\paragraph{Common benchmarks} For a more extensive comparison with previous work, we include results on English--$\{$Finnish, German, Italian$\}$ dictionaries used in \citet{Conneau2018} and \newcite{Artetxe2018}---the second best approach to BDI known to us, which also uses Procrustes Analysis. We conduct experiments using three forms of supervision: gold-standard seed dictionaries of 5000 word pairs, cross-lingual homographs, and numerals. We use train and test bilingual dictionaries from \citet{Dinu2015} for English-Italian and from \citet{Artetxe2017} for English-$\{$Finnish, German$\}$. %\footnote{The lexicons similarly to before consist of training sets of 5000 unique word pairs and test sets of 1500 unique word pairs.}
Following \citet{Conneau2018}, we report results with a set of CBOW embeddings trained on the WaCky corpus \citep{Barone2016}, and with Wikipedia embeddings. 

\begin{table*}
\centering
\resizebox{\linewidth}{!}{
\begin{tabular}{l|ccc|ccc|ccc}%|ll|ll}
\toprule
&\multicolumn{3}{c|}{\sc it}&\multicolumn{3}{c|}{\sc de}&\multicolumn{3}{c}{\sc fi}\\
&\multicolumn{1}{c}{5000}&\multicolumn{1}{c}{Identical}&\multicolumn{1}{c|}{Numerals}&\multicolumn{1}{c}{5000}&\multicolumn{1}{c}{Identical}&\multicolumn{1}{c|}{Numerals}&\multicolumn{1}{c}{5000}&\multicolumn{1}{c}{Identical}&\multicolumn{1}{c}{Numerals}\\
\midrule
&\multicolumn{9}{c}{\sc Wacky}\\
\midrule
\multirow{1}{*}{\newcite{Artetxe2018}}&45.27*&38.33&39.40*&44.27*&40.73&40.27*&32.94*&27.39&26.47*\\% &&&&\\
\midrule
\multirow{1}{*}{PA}&44.90&45.47&01.13&47.26&47.20&45.93&{\bf 33.50}&\textbf{31.46}&01.05 \\% &&&&\\
\midrule
\multirow{1}{*}{GPA}&{\bf 45.33}&\textbf{45.80}&{\bf 45.93} %(27e)
&{\bf 48.46}&\textbf{47.60}&{\bf 47.60}&31.39&31.04&{\bf 28.93} \\%(74e) \\%
\midrule
&\multicolumn{9}{c}{\sc Wikipedia}\\
\midrule
\multirow{1}{*}{PA}& 66.24&66.39&-&65.33&64.77&-&36.77&35.40&- \\%
\midrule
\multirow{1}{*}{GPA}&\textbf{67.60}&\textbf{67.14}&-&{\bf 66.21}&\textbf{65.81}&-&{\bf 38.14}&\textbf{37.87}&-\\ % is GPA 67.6(0) - it missed a decimal? --> Yes
\bottomrule
\end{tabular}
}
%\multirow{1}{*}{PA}& 77.47&77.73&-&74.27&73.67&-&\textbf{47.60}&45.33&- \\%
%\midrule
%\multirow{1}{*}{GPA}&\textbf{78.27}&\textbf{78.07}&-&\textbf{74.53}&\textbf{74.40}&-&47.40&\textbf{45.27}&-\\ 
\caption{Results on standard benchmarks, measured in P@1. \** Results as reported in the original paper. {\bf Notes:} \newcite{Conneau2018} report 63.7 on Italian with Wikipedia embeddings; results with different embedding sets are not comparable due to a non-zero out-of-vocabulary rate on the test set for Wikipedia embeddings; Wikipedia embeddings are trained on corpora with removed numerals, so supervision from numerals cannot be applied. } 
\label{tab:standard-benchmarks}
\end{table*}

\textbf{Results} are reported in Table \ref{tab:standard-benchmarks}. We observe that GPA outperforms PA consistently on Italian and German with the WaCky embeddings, and on all languages with the Wikipedia embeddings. Notice that once more, Finnish benefits the most from a switch to GPA in the Wikipedia embeddings setting, but it is also the only language to suffer from that switch in the WaCky setup. 

Interestingly, PA fails to learn a good alignment for Italian and Finnish when supervised with numerals, while GPA performs comparably with numerals as with other forms of supervision. 
%It is with supervision from numerals, actually that GPA obtains the highest score with WaCky embeddings for Italian: 44.93. 
\citet{Conneau2018} point out that improvement from subsequent iterations of PA is generally negligible, which we also found to be the case. We also found that while PA learned a slightly poorer alignment than GPA, it did so faster. With our criterion for early stopping, PA converged in 5 to 10 epochs, while GPA did so within 10 to 15 epochs\footnote{Notice that one epoch with both PA and GPA takes less than half a minute, so the slower convergence of GPA is in no way prohibitive.}
. In the case of Italian and Finnish alignment supervised by numerals, PA converged in 8 and 5 epochs, respectively, but clearly got stuck in local minima. GPA took considerably longer to converge: 27 and 74 epochs, respectively, but also managed to find a reasonable alignment between the language spaces. This points to an important difference in the learning properties of PA and GPA---unlike PA, GPA has a two-fold objective of opposing forces: it is simultaneously aligning each embedding space to two others, thus pulling it in different directions. This characteristic helps GPA avoid particularly adverse local minima. 

\subsection{Multi-support GPA}
In these experiments, we perform GPA with $k=3$, including a third, linguistically-related supporting language in the alignment process. To best evaluate the benefits of the multi-support setup, we use as targets five low-resource languages: Afrikaans, Bosnian, Estonian, Hebrew and Occitan (see statistics in Table~\ref{tab:stats})\footnote{Occitan dictionaries were not available from the MUSE project, so we extracted a test dictionary of 911 unique word pairs from an English-Occitan lexicon available at \url{http://www.occitania.online.fr/aqui.comenca.occitania/en-oc.html}.}.  Three-way dictionaries, both the initial one (consisting of cross-lingual homographs) and subsequent ones, are obtained by assuming transitivity between two-way dictionaries: if two pairs of words, $e^m$--$e^n$ and $e^m$--$e^l$, are deemed translational pairs, then we consider $e^n$--$e^m$--$e^l$ a translational triple. 

\begin{table*}
\resizebox{\textwidth}{!}{
\begin{tabular}{l|ll|ll|ll|ll|ll|ll}
\toprule
&\multicolumn{2}{c|}{\sc af}&\multicolumn{2}{c|}{\sc bs}&\multicolumn{2}{c|}{\sc et}&\multicolumn{2}{c|}{\sc he}&\multicolumn{2}{c|}{\sc oc}&\multicolumn{2}{c}{Ave}\\
& $k=1$ & $k=10$& $k=1$ & $k=10$& $k=1$ & $k=10$& $k=1$ & $k=10$& $k=1$ & $k=10$& $k=1$ & $k=10$\\
\midrule
PA&28.87&50.53&22.40&48.40&30.00&57.93&37.53&67.27&17.12&33.26&27.18&51.48\\
GPA&{\bf 29.93}&\textbf{50.67}&{\bf 24.20}&\textbf{50.20}&\textbf{31.87}&\textbf{60.07}&38.93&\textbf{68.93}&17.12&34.91&28.41&52.96\\
MGPA&28.93&49.20&21.00&48.60&30.73&59.53&37.53&66.47&{\bf 23.82}&\textbf{40.18}&28.40&52.80\\
MGPA$^+$&28.80&49.20&23.46&48.87&31.27&59.80&{\bf 40.40}&68.80&22.83&38.53&{\bf 29.35}&\textbf{53.04}\\
\bottomrule
\end{tabular} }
\caption{Results for low-resource languages with PA, GPA and two multi-support settings.}
\label{tab:low-resource}
\end{table*}

We report \textbf{results} in Table \ref{tab:low-resource} with multi-support GPA in two settings: a three-way alignment trained for 10 epochs (MGPA), and a three-way alignment trained for 10 epochs, followed by 5 epochs of two-way fine-tuning (MGPA+).  
%\paragraph{Results} 
We observe that at least one of our new methods always improves over PA. GPA always outperforms PA and it also outperforms the multi-support settings on three out of five languages. Yet, results for Hebrew and especially for Occitan, are best in a multi-support setting---we thus mostly focus on these two languages in the following subsections. 

\begin{figure}
\centering
\subfloat[Occitan]{%
  \includegraphics[clip,width=0.7\columnwidth]{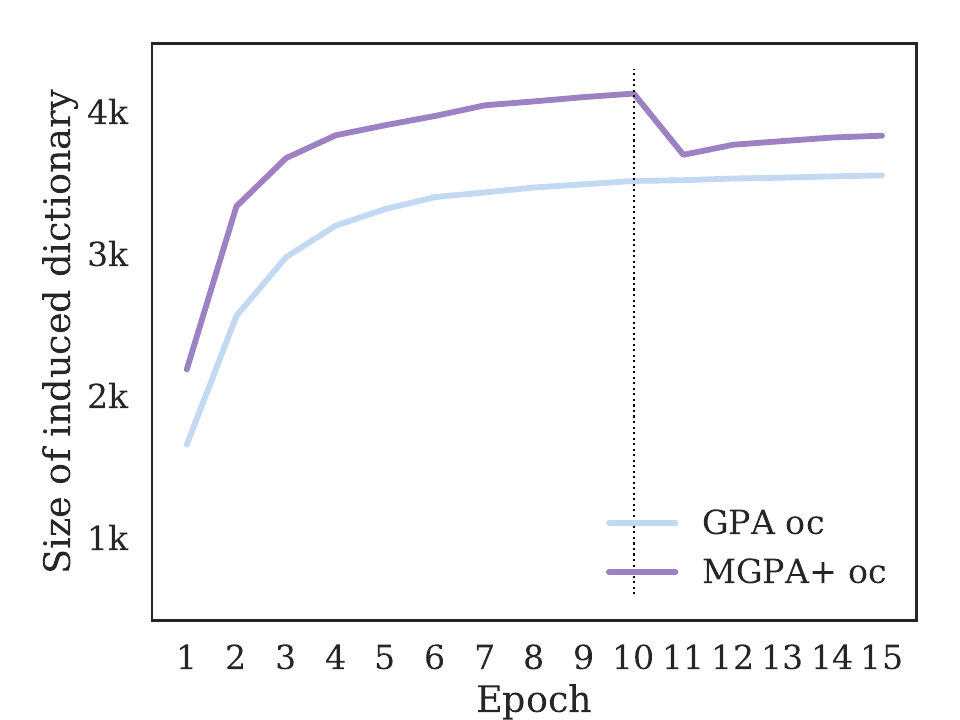}%
}

\subfloat[Hebrew]{%
  \includegraphics[clip,width=0.7\columnwidth]{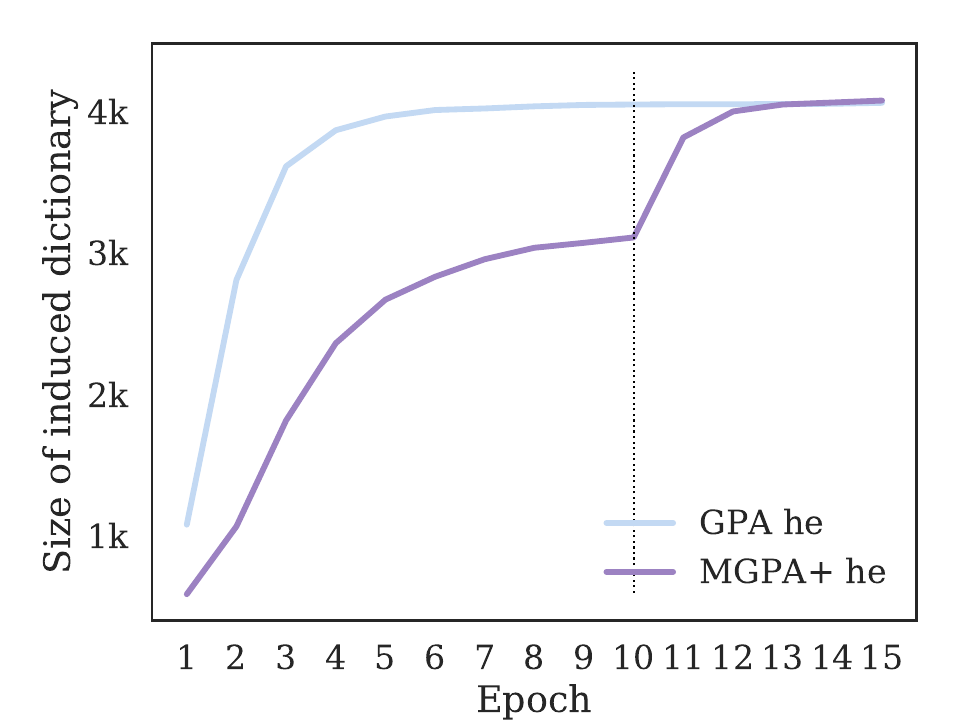}%
}
\caption{Progression of dictionary size during GPA and MGPA+ training. The dotted line marks the boundary between MGPA and fine-tuning.}
\label{dict_size}
\end{figure}

\paragraph{MGPA} has variable performance: for four languages precision suffers from the addition of a third language, e.g. compare 38.93 for Hebrew with GPA to 37.53 with MGPA; for Occitan, however, the most challenging target language in our experiments, MGPA beats all other approaches by a large margin: 17.12 with GPA versus 23.81 with MGPA. This pattern relates to the effect a supporting language has on the size of the induced seed dictionary. Figure~\ref{dict_size} visualizes the progression of dictionary size during training with and without a supporting language for Occitan and Hebrew. The portion of the purple curves to the left of the dotted line corresponds to MGPA: notice how the curves are swapped between the two plots. 
Spanish\textit{ actually} provides support for the English-Occitan alignment, by contributing to an increasingly larger seed dictionary---this provides better anchoring for the learned alignment.
Having Arabic as support for English-Hebrew alignment, on the other hand, causes a considerable reduction in the size of the seed dictionaries, giving GPA less anchor points and thus damaging the learned alignment. % (precision for Hebrew is higher with GPA than with MGPA). 
The variable effect of a supporting language on dictionary size, and consequently on alignment precision, relates to the quality of alignment of the support language with English and with the target language: referring back to Table~\ref{tab:pa-gpa-comparison}, English-Spanish, for example, scores at 81.93, while English-Arabic precision is 35.33. Notice that despite our linguistically-motivated choice to pair related low- and high-resource languages for multi-support training, it is not necessarily the case that those should align especially well, as that would also depend on practical factors, such as embeddings quality and training corpora similarity \cite{Søgaard2018}.   
%\paragraph{MGPA+}builds off of MGPA, and performs additional two-way fine-tuning. This makes for 

\paragraph{MGPA+} applies two-way fine-tuning on top of MGPA. This leads to a drop in precision for Occitan, due to the removed support of Spanish and the consequent reduction in size of the induced dictionary (observe the fall of the purple curve after the dotted line in Figure ~\ref{dict_size} (a)). Meanwhile, precision for Hebrew is highest with MGPA+ out of all methods included. While Arabic itself is not a good support language, its presence in the three-way MGPA alignment seems to have resulted in a good initialization for the English-Hebrew two-way fine-tuning, thus helping the model reach an even better minimum along the loss curve.

\section{Discussion: Why it works} If word vector spaces were completely isomorphic, the introduction of a third (or fourth) space, and the application of GPA, would lead to the same alignment as the alignment learned by PA, projecting the source language $E$ into the target space $F$. This follows from the transitivity of isomorphism: if $E$ is isomorphic to $G$ and $G$ is isomorphic to $F$, then $E$ is isomorphic to $F$, via the isomorphism obtained by composing the isomorphisms from $E$ to $G$ and from $G$ to $F$. 
So why do we observe improvements?

\citet{Søgaard2018} have shown that word vector spaces are often relatively far from being isomorphic, and approximate isomorphism is not transitive. What we observe therefore appears to be an instance of the Poincar\'{e} Paradox \cite{Poincare:02}. %Poincar\'{e} points out that a flame is similar to the moon because they both appear luminous, and the moon is similar to a ball because they both are round; a flame and a ball, however, are very dissimilar. The algorithm we propose here works, so to speak, by hallucinating a moon (a vector space $C$) to align a flame ($A$) and a ball ($B$). While it may be hard for flames and balls to live in the same space, the pairwise alignments $A-C$ and $B-C$ are possible, which is better than randomly aligning $A$ to $B$ in the blind. 
While GPA is not more expressive than PA, it may still be easier to align each monolingual space to an intermediate space, as the latter constitutes a more similar target (albeit a non-isomorphic one); for example, the loss landscape of aligning a source and target language word embedding with an average of the two may be much smoother than when aligning source directly with target.  Our work is in this way similar in spirit to \newcite{Raiko:ea:12}, who use simple linear transforms to make learning of non-linear problems easier. 

\begin{table*}
\centering
\begin{tabular}{c|l|l|l|l|l}
\toprule
&{\sc query}&{\sc gold}&{\sc PA}&{\sc GPA}&{\sc MGPA+}\\
\midrule
 \parbox[t]{2mm}{\multirow{16}{*}{\rotatebox[origin=c]{90}{PA \xmark,   GPA \cmark }}}&variraju&vary&varies&vary&varies\\  
&kanjon&canyon&headwaters&canyon&headwaters\\ 
&dijalog&dialogue&dialogues&dialogue&dialogue\\ 
&izjava&statement&deniable&statement&statements\\ 
&plazme&plasma&conduction&plasma&microspheres\\ 
&računari&computers&minicomputers&computers&mainframes\\ 
&aparat&apparatus&duplex&apparatus&apparatus\\ 
&sazviježđa&constellations&asterisms&constellations&constellations\\ 
&uspostavljanje&establishing&reestablishing&establishing&establishing\\ 
&industrijska&industrial&industry&industrial&industrial\\ 
&stabilna&stable&unstable&stable&stable\\ 
&disertaciju&dissertation&habilitation&dissertation&thesis\\ 
&protivnici&opponents&opposing&opponents&opponents\\ 
&pozitivni&positive&negative&positive&positive\\ 
&instalacija&installation&installations&installation&installation\\ 
&duhana&tobacco&liquors&tobacco&tobacco\\ 
\midrule
 \parbox[t]{2mm}{\multirow{17}{*}{\rotatebox[origin=c]{90}{PA \cmark,   GPA \xmark }}}&hor&choir&choir&musicum&choir\\ 
&crijevo&intestine&intestine&intestines&intestine\\ 
&vidljiva&visible&visible&unnoticeable&visible\\ 
&temelja&foundations&foundations&superstructures&pillars\\ 
&kolonijalne&colonial&colonial&colonialists&colonialists\\ 
&spajanje&merger&merger&merging&merging\\ 
&suha&dry&dry&humid&dry\\ 
&janez&janez&janez&mariza&janez\\ 
&kampanju&campaign&campaign&campaigning&campaign\\ 
&migracije&migration&migration&migrations&migrations\\ 
&sobu&room&room&bathroom&bathroom\\ 
&predgrađu&suburb&suburb&outskirts&suburb\\ 
&specijalno&specially&specially&specialist&specially\\ 
&hiv&hiv&hiv&meningococcal&hiv\\ 
&otkrije&discover&discover&discovers&discover\\ 
&proizlazi&arises&arises&differentiates&deriving\\ 
&tajno&secretly&secretly&confidentially&secretly\\ 
\midrule
 \parbox[t]{2mm}{\multirow{14}{*}{\rotatebox[origin=c]{90}{PA \xmark,   GPA \xmark }}}&odred&squad&reconnoitre&stragglers&skirmished\\ 
&učesnik&attendee&participant&participant&participant\\ 
&saznao&learned&confided&confided&confided\\ 
&dobiva&gets&earns&earns&earns\\ 
&harris&harris&guinn&zachary&zachary\\ 
&snimke&videos&footage&footages&footage\\
&usne&lips&ear&ear&toes\\ 
&ukinuta&lifted&abolished&abolished&abolished\\ 
&objave&posts&publish&publish&publish\\ 
&obilježje&landmark&commemorates&commemorates&commemorates\\ 
&molim&please&appologize&thank&kindly\\ 
&čvrste&solid&concretes&concretes&concretes\\ 
&intel&intel&genesys&motorola&transputer\\ 
&transformacije&transformations&transformation&transformation&transformation\\ 
\end{tabular}
\caption{Example translations from Bosnian into English.}
\label{tab:examples}
\end{table*}

\subsection{Error Analysis}
Table~\ref{tab:examples} lists example translational pairs as induced from alignments between English and Bosnian, learned with PA, GPA and MGPA+. For interpretability, we query the system with words in Bosnian and seek their nearest neighbors in the English embedding space. P@1 over the Bosnian-English test set of \citet{Conneau2018} is 31.33, 34.80, and 34.47 for PA, GPA and MGPA+, respectively. The examples are grouped in three blocks, based on success and failure of PA and GPA alignments to retrieve a valid translation.

It appears that a lot of the difference in performance between PA and GPA concerns \textbf{morphologically related words}, e.g. \textit{campaign} v. \textit{campaigning}, \textit{dialogue }v.\textit{ dialogues}, \textit{merger} v. \textit{merging} etc. These word pairs are naturally confusing to a BDI system, due to their related meaning and possibly identical syntactic properties (e.g. \textit{merger} and \textit{merging} can both be nouns). Another common mistake we observed in mismatches between PA and GPA predictions, was the wrong choice between two \textbf{antonyms}, e.g. \textit{stable} v. \textit{unstable} and \textit{visible} v. \textit{unnoticeable}.  Distributional word representations are known to suffer from limitations with respect to capturing opposition of meaning \citep{Mohammad13}, so it is not surprising that both PA- and GPA-learned alignments can fail in making this distinction. While it is not the case that GPA always outperforms PA on a query-to-query basis in these rather challenging cases, on average GPA appears to learn an alignment more robust to subtle morphological and semantic differences between neighboring words. Still, there are cases where PA and GPA both choose the wrong morphological variant of an otherwise correctly identified target word, e.g. \textit{transformation} v. \textit{transformations}. 

Notice that many of the queries for which both algorithms fail, do result in a \textbf{nearly synonymous word} being predicted, e.g. \textit{participant} for \textit{attendee}, \textit{earns} for \textit{gets}, \textit{footage} for \textit{video}, etc. This serves to show that the learned alignments are generally good, but they are not sufficiently precise. This issue can have two sources: a suboptimal method for learning the alignment and/or a ceiling effect on how good of an alignment can be obtained, within the space of orthogonal linear transformations. 
\subsection{Procrustes fit}

To explore the latter issue and to further compare the capabilities of PA and GPA, we perform a \textit{Procrustes fit} test, where we learn alignments in a fully supervised fashion, using the test dictionaries of \citet{Conneau2018}\footnote{For Occitan, we use our own test dictionary.} for both training \textit{and} evaluation\footnote{In this experiment, we only run a single epoch of each alignment algorithm, as that is guaranteed to give us the best Procrustes fit for the particular set of training word pairs we would then evaluate on.}. In the ideal case, i.e. if the subspaces defined by the words in the seed dictionaries are perfectly alignable, this setup should result in precision of 100\%. 

\begin{figure*}
\includegraphics[clip,width=\linewidth]{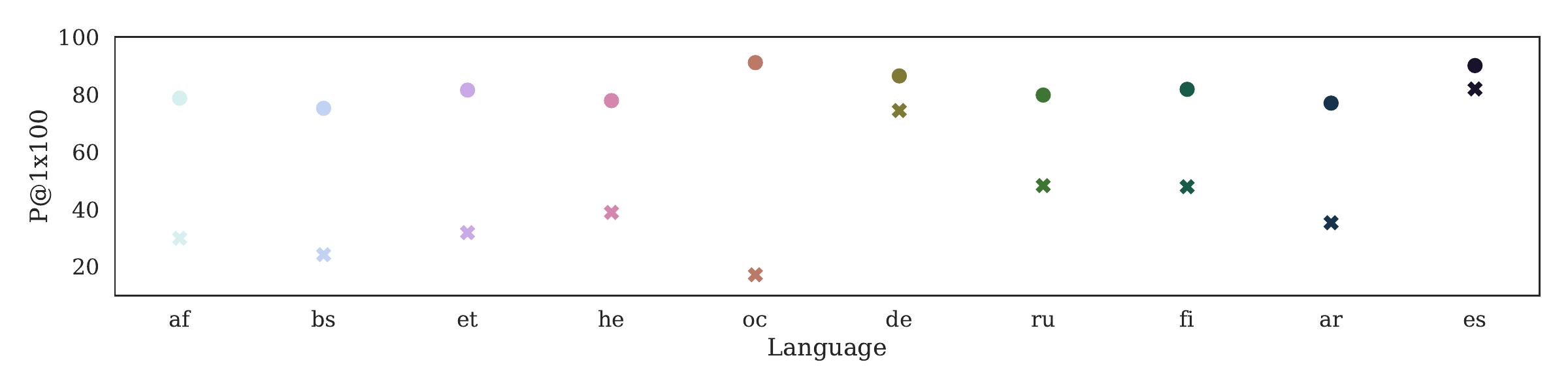}
\caption{Procrustes fit test. Circles mark the results from fitting and evaluating GPA on the test dictionaries to measure the \textit{Procrustes fit}. \textbf{x}s mark the weakly-supervised results reported in Tables~\ref{tab:pa-gpa-comparison} and \ref{tab:low-resource}.}
\label{fig:profit}
\end{figure*}

We found the difference between the fit with PA and GPA to be negligible, 0.20 on average across all 10 languages (5 low-resource and 5 high-source languages). It is not surprising that PA and GPA results in almost equivalent fits---the two algorithms both rely on linear transformations, i.e. they are equal in expressivity. As pointed out earlier, the superiority of GPA over PA stems from its more robust learning procedure, not from higher expressivity. Figure~\ref{fig:profit} thus only visualizes the Procrustes fit as obtained with GPA. 

The Procrustes fit of all languages is indeed lower than 100\%, showing that there is a \textbf{ceiling on the linear alignability} between the source and target spaces. We attribute this ceiling effect to variable degrees of linguistic difference between source and target language and possibly to differences in the contents of cross-lingual Wikipedias (recall that the embeddings we use are trained on Wikipedia corpora). An apparent correlation emerges between the Procrustes fit and precision scores for weakly-supervised GPA, i.e. between the circles and the \textbf{x}s in the plot. The only language that does not conform here is Occitan, which has the highest Procrustes fit and the lowest GPA precision out of all languages, but this result has an important caveat: our dictionary for Occitan comes from a different source and is much smaller than all the other dictionaries.

For some of the high-resource languages, weakly-supervised GPA takes us rather close to the best possible fit: e.g. for Spanish GPA scores 81.93\%, and the Procrustes fit is 90.07\%. While low-resource languages do not necessarily have lower Procrustes fits than high-resource ones (compare Estonian and Finnish, for example), the gap between the Procrustes fit and GPA precision is on average much higher within low-resource languages than within high-resource ones (52.46\footnote{Even if we leave Occitan out as an outlier, this number is still rather high: 47.10.} compared to 25.47, respectively). This finding is in line with the common understanding that the quality of distributional word vectors depends on the amount of data available---we can infer from these results that suboptimal embeddings results in suboptimal cross-lingual alignments. 
\subsection{Multilinguality}
Finally, we note that there may be specific advantages to including support languages for which large monolingual corpora exist, as those should, theoretically, be easier to align with English (also a high-resource language): variance in vector directionality, as studied in \citet{Mimno:Thompson:17},  increases with corpus size, so we would expect embedding spaces learned from corpora comparable in size, to also be more similar in shape.

\section{Related work}

\paragraph{Bilingual embeddings} Many diverse cross-lingual word embedding models have been proposed \cite{Ruder2018}. The most popular kind learns a linear transformation from source to target language space \cite{Mikolov2013e}. In most recent work, this mapping is constrained to be orthogonal and solved using Procrustes Analysis \cite{Xing2015,Artetxe2017,Artetxe2018,Conneau2018,Lu2015a}. The approach most similar to ours,  \citet{Faruqui2014}, uses canonical correlation analysis (CCA) to project both source and target language spaces into a third, joint space. In this setup, similarly to GPA, the third space is iteratively updated, such that at timestep $t$, it is a product of the two language spaces as transformed by the mapping learned at timestep $t-1$. The objective that drives the updates of the mapping matrices is to maximize the correlation between the projected embeddings of translational equivalents (where the latter are taken from a gold-standard seed dictionary). In their analysis of the transformed embedding spaces, \citet{Faruqui2014} focus on the improved quality of monolingual embedding spaces themselves and do not perform evaluation of the task of BDI. They find that the transformed monolingual spaces better encode the difference between synonyms and antonyms: in the original monolingual English space, synonyms and antonyms of \textit{beautiful} are all mapped close to each other in a mixed fashion; in the transformed space the synonyms of \textit{beautiful} are mapped in a cluster around the query word and its antonyms are mapped in a separate cluster. This finding is in line with our observation that GPA-learned alignments are more precise in distinguishing between synonyms and antonyms.

\paragraph{Multilingual embeddings} Several approaches extend existing methods to space alignments between more than two languages \cite{Ammar2016a,Ruder2018}. \citet{Smith2017} project all vocabularies into the English space. In some cases, multilingual training has been shown to lead to improvements over bilingually trained embedding spaces \cite{Vulic2017}, similar to our findings.

\section{Conclusion} Generalized Procrustes Analysis yields benefits over simple Procrustes Analysis for Bilingual Dictionary Induction, due to its smoother loss landscape. In line with earlier research, benefits from the introduction of a common latent space seem to relate to a better distinction of synonyms and antonyms, and of syntactically-related words. GPA also offers the possibility to include multi-lingual support for inducing a larger seed dictionary during training, which better anchors the English to target language alignment in low-resource scenarios.  

\section*{Acknowledgements}

Sebastian is supported by Irish Research Council Grant Number EBPPG/2014/30 and Science Foundation Ireland Grant Number SFI/12/RC/2289, co-funded by the European Regional Development Fund.

\bibliography{general_procrustes}
\bibliographystyle{acl_natbib_nourl}

\end{document}